\documentclass[titlepage,12pt]{article}

\usepackage[citestyle=authoryear,backend=biber,maxcitenames=1]{biblatex}
\newcommand{\citep}{\autocite}
\newcommand{\citet}{\textcite}

\usepackage[margin=1in]{geometry}
\usepackage{graphicx}
\usepackage{indentfirst}
\edef\restoreparindent{\parindent=\the\parindent\relax}
\usepackage[indent]{parskip}
\restoreparindent

\usepackage{setspace}
\usepackage{commath}

\usepackage{xcolor}

\usepackage{hyperref}

\usepackage{amssymb}
\usepackage{mathrsfs}

\usepackage{graphicx}
\usepackage{subcaption}

\usepackage{xurl}

\title{Boulder2Vec: Modeling Climber Performances in Professional Bouldering Competitions}
\author{Ethan Baron, Victor Hau, Zeke Weng\\ \\University of Toronto Sports Analytics Student Group}
\date{September 2024}

\begin{document}

\maketitle

\onehalfspacing
\newpage

\section{Introduction}
In the past decade, sport climbing has grown to be a popular pastime due to its social, physical and mental stimulation. This growth has been bolstered by its recent addition to the Summer Olympics in three formats: bouldering, speed and lead. In particular, bouldering, a form of climbing that focuses on short, difficult routes known as ``problems”, has seen the greatest growth, with 71\% of new climbing gyms opening in North America being boulder-focused \citep{climbing_business_journal}.

Using professional bouldering competition data from 2007 to 2022, we train a logistic regression to predict climber results and measure climber skill. However, this approach is limited, as a single numeric coefficient per climber cannot adequately capture the intricacies of each climbers’ varying strengths and weaknesses in different boulder problems. For example, some climbers might prefer more static, technical routes while other climbers may specialize in powerful, dynamic routes.

To this end, we apply Probabilistic Matrix Factorization (PMF), a framework commonly used in recommender systems, to represent the unique characteristics of climbers and problems with latent, multi-dimensional vectors. In this framework, a climber’s performance on a given problem is predicted by taking the dot product of the corresponding climber vector and problem vectors. PMF effectively handles sparse datasets, such as our dataset, where only a subset of climbers attempt each particular problem, by extrapolating patterns from similar climbers.

We contrast the empirical performance of PMF to the logistic regression approach and investigate the multivariate representations produced by PMF to gain insights into climber characteristics. Our results show that the multivariate PMF representations improve predictive performance of professional bouldering competitions by capturing both the overall strength of climbers and their specialized skill sets in interpretable embeddings. We provide our code open-source at \url{https://github.com/baronet2/boulder2vec}.

\section{Background}

Professional bouldering competitions consist of a series of problems. Climbers are given a set number of minutes to attempt to reach the ``top'' hold of the problem, scoring the maximum number of points, or an intermediate ``zone'' hold, scoring fewer points. If the climber falls before reaching the top hold, they may continue attempting the problem until the time runs out. Finally, the climbers are ranked by their combined score across all the problems.

One interesting aspect of bouldering competitions is that the layout of each problem is uniquely designed for that competition, and unknown to the climbers prior to the competition. Different problems can require distinct techniques or strengths to be completed. Prior work has been done in categorizing these problems. For example, \citet{boulder_types} label five types of problems according to the technique required to complete the most difficult section of the problem: Dynamo, Volume, Crimp, Slab, and Mantle.

\subsection{Relevant Prior Work}

The performance of professional climbers in bouldering competitions has not been studied extensively. \citet{bouldering_analysis} collects a dataset of results from professional bouldering competitions between 2007 and 2022 and performs some exploratory analysis of the data, but does not develop a model for climber performance. \citet{bayesian_bouldering} builds a Bayesian logistic regression to predict performance on bouldering problems, but they focus on a dataset of amateur climbers attempting outdoor bouldering routes rather than professional bouldering competitions. \citet{boulder_types} do analyze performances of athletes in professional bouldering competitions but aggregate performance by gender rather than modelling the skill of individual climbers. Therefore, this work presents a novel statistical model of climber performance in professional bouldering competitions.

While we do consider a logistic regression approach in the vein of \citet{bayesian_bouldering}, this approach only extracts a single skill rating for each climber. However, boulder problems can include a variety of different features, therefore demanding a variety of different skills from the climbers. 
Just as \citet{boulder_types} find statistically significant differences in performance between elite (top 20) and non-elite climbers on different boulder types, we hypothesize that individual climbers might also have strengths in particular areas. For example, a specific climber might excel at dynamic problems but be less successful at technical slab problems. Therefore, we seek a model that allows us to capture the particular skill set of climbers, rather than a unilateral strength rating.

Clearly, such a model must learn a multi-dimensional representation for climber skill. While such a model has not yet been proposed for bouldering, there have been prior works applying this idea to other sports. For example, \citet{tennis} extends the classic Elo rating approach to learn correlated latent skills for tennis players on clay, grass, and hard court surfaces. A similar approach could be used to learn climber strength ratings for each 
boulder category, as identified by \citet{boulder_types}. However, rather than limit ourselves to this simple discrete categorization of problems which must be labelled manually, we seek an automated approach that can learn a more nuanced, dense representation of boulder problem. This closely resembles the work of \citet{bike2vec}, which learns multi-dimensional representations for professional road cycling athletes and races.


\section{Methods}

\subsection{Dataset}

\begin{figure}[htbp]
    \centering
    \begin{subfigure}[b]{0.56\textwidth}
        \centering
        \includegraphics[width=\linewidth]{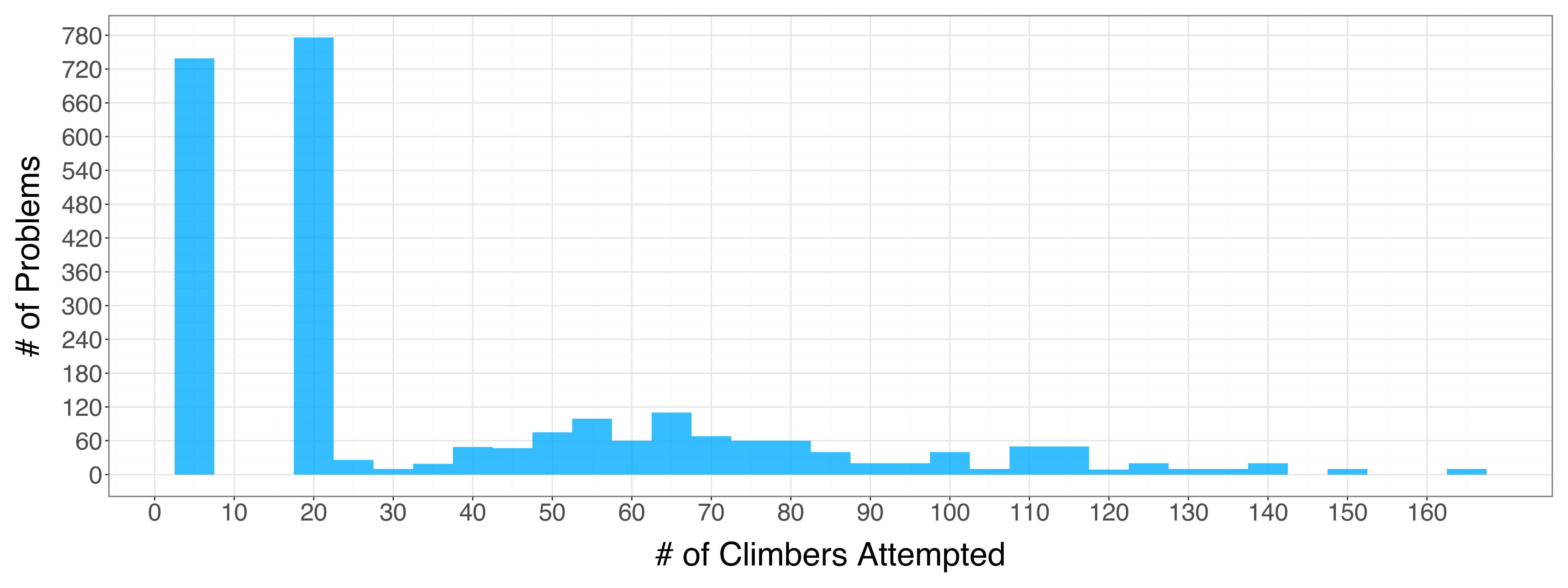}
        \caption{}
        \label{fig:problem-id-cdf}
    \end{subfigure}
    \hfill
    \begin{subfigure}[b]{0.425\textwidth}
        \centering
        \includegraphics[width=\linewidth]{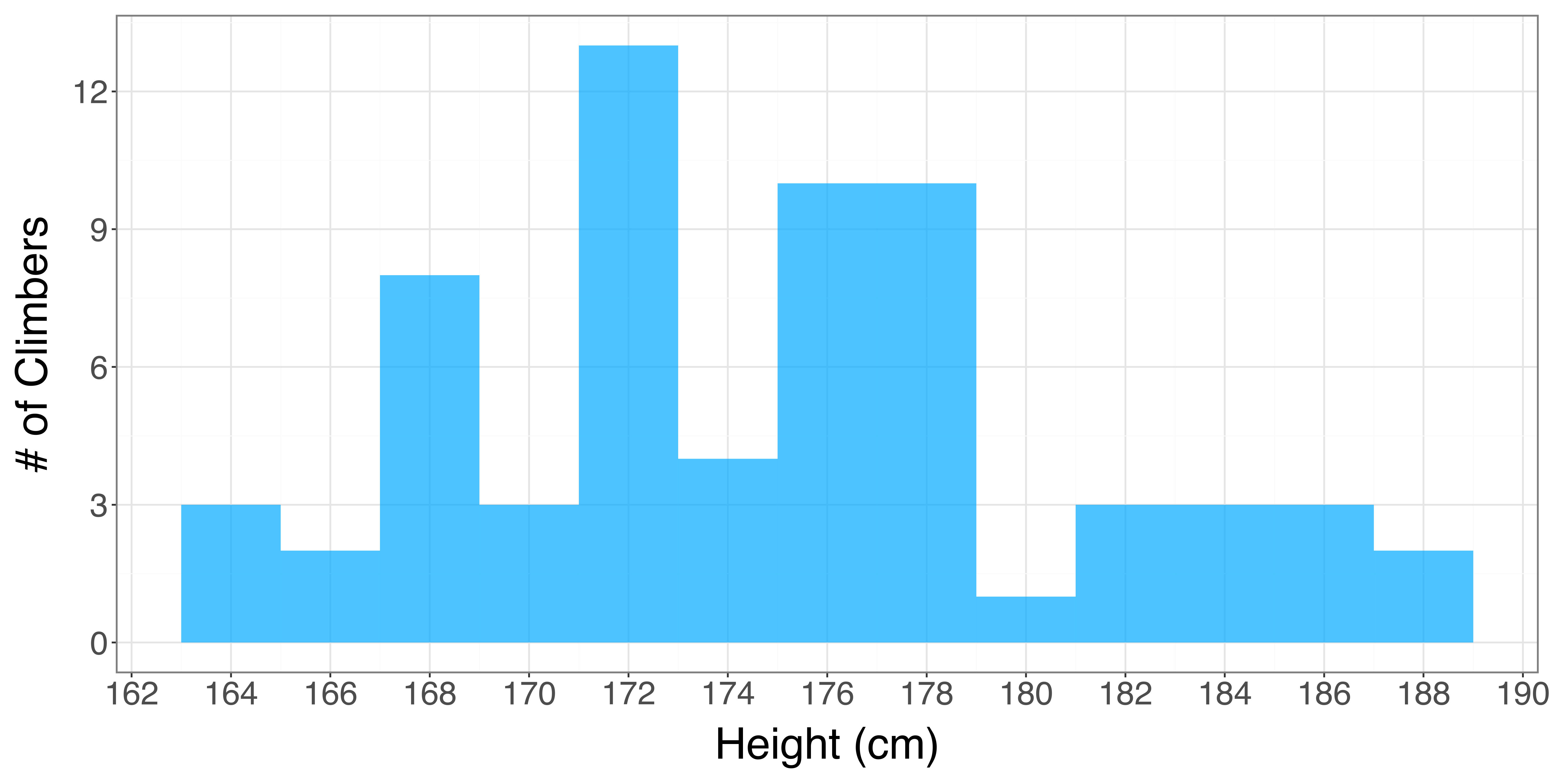}
        \caption{}
        \label{fig:heights}
    \end{subfigure}
    \caption{Distribution of number of climbers per problem (a) and climber heights (b).}
    \label{fig:eda}
\end{figure}

We use the publicly available dataset gathered by \citet{bouldering_analysis}, which includes results from all men's bouldering events hosted by the International Federation of Sport Climbing (IFSC) from 2007 to 2022. This consists of almost 380,000 attempts across 2517 unique problems in 101 competitions during that timespan. These problems range across the various rounds of competition, from qualifier stages to finals. In our work, we consider reaching the ``top'' and ``zone'' holds on a given problem to be two distinct boulder problems. This essentially doubles the amount of data available, although we acknowledge that performance on the ``top'' and ``zone'' are correlated in a given problem. The distribution of the number of climbers attempting each problem in our dataset is shown in Figure \ref{fig:problem-id-cdf}.

We further augment our dataset by collecting data on each climber's height from the official IFSC website, where available. A histogram of the distribution of available climber heights is presented in Figure \ref{fig:heights}. 

%

    
\subsection{Logistic Regression}

As a baseline model, we use a basic logistic regression without any regularization penalty. Specifically, the probability that climber $i$ successfully completes problem $j$ of type $t$ in round $r$ is:
\begin{equation}
    P(y_{ij} = 1) = \sigma(\beta_0 + \beta_r + \beta_t + \beta_i)
\end{equation}
where $\sigma(x) = \frac{1}{1 + e^{-x}}$ is the logistic function. We include a round (qualifier/semi-final/final) coefficient $\beta_r$,  problem type (top/zone) coefficient $\beta_t$, and a global intercept $\beta_0$ as the predictors for problem difficulty. These are one-hot encoded and fed into the model. Most importantly, we include a climber-specific effect $\beta_i$ which measures the climber's overall skill. Under this framework, the most skilled climbers will have the highest coefficients $\beta_i$ since this maximizes their predicted probability of success.

\begin{figure}[h!]
    \centering
    \includegraphics[width=\textwidth,height=\textheight,keepaspectratio]{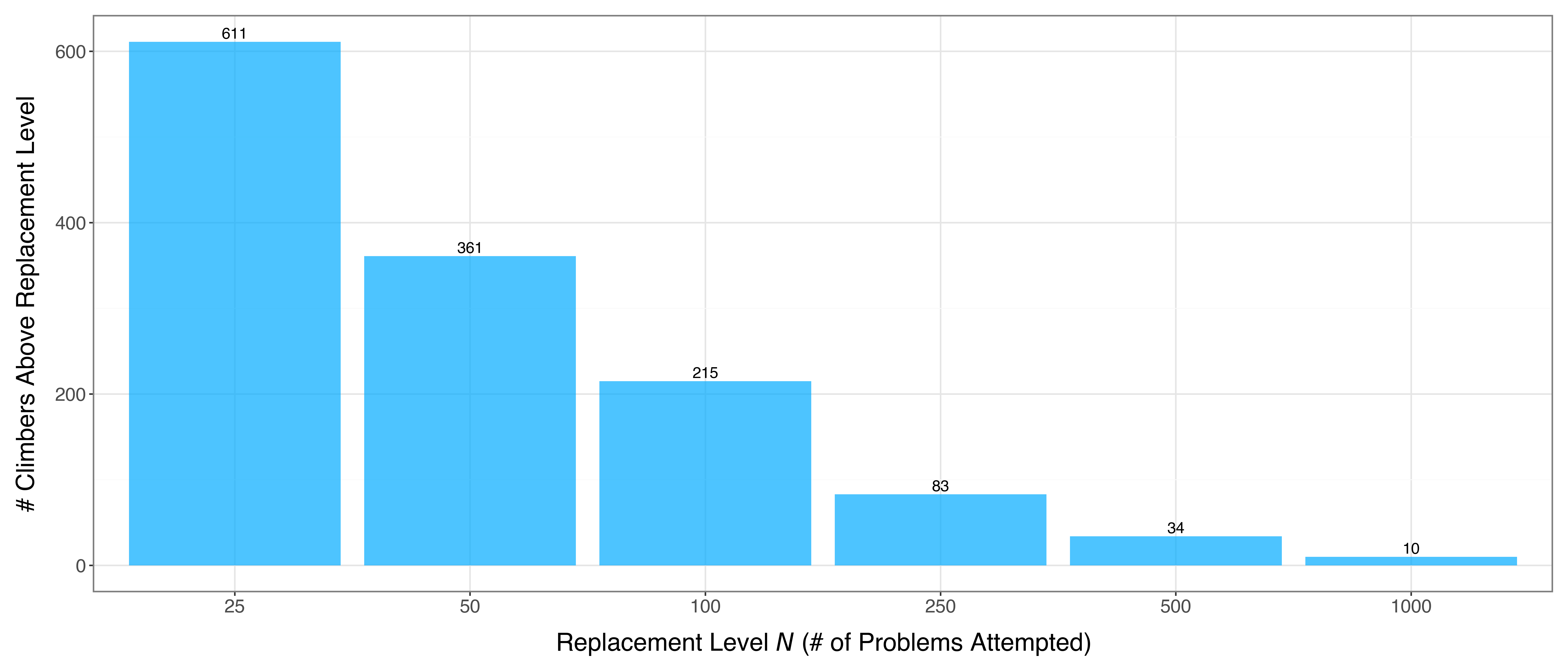}
    \caption{Number of climbers with sufficient data at each replacement level $N$.}
    \label{fig:climbers_above_replacement_level}
\end{figure}

To prevent overfitting, we only learn a climber-specific coefficient for climbers with at least $N$ problems attempted in the training set. We call this hyperparameter $N$ the ``replacement level'', and experiment with the settings $N \in \{25, 50, 100, 250, 500, 1000\}$. Any climbers with fewer than $N$ problems attempted in the training set are grouped together and treated as one ``replacement-level climber''. The number of climbers not grouped together at each replacement level $N$ is shown in Figure \ref{fig:climbers_above_replacement_level}. We use 5-fold testing of the model to assess its generalizability and quantify uncertainty in the model's performance.

\subsection{Probabilistic Matrix Factorization}

To extract multi-dimensional representations of climber skill, we employ Probabilistic Matrix Factorization (PMF) \citep{pmf}. Let the matrix $Y \in \mathbb{R} ^ {m \times n}$ represent the binary observed outcomes, where the element $Y_{ij}$ indicates whether climber $i$ succeeded to complete problem $j$. Here, $m$ is the number of climbers in the dataset, and $n$ is the number of problems in the dataset.

Note that the matrix $Y$ is sparse, since for each given problem only a subset of climbers in the dataset (namely those participating in the respective competition), will have recorded attempts. This sparsity necessitates the use of the PMF framework rather than a simpler matrix factorization algorithm such as non-negative matrix factorization (NNMF).

In the PMF framework, we seek to learn matrices $U \in \mathbb{R}^{m \times d}$ and $V \in \mathbb{R} ^ {d \times n}$ such that $UV \approx Y$. Each row of the matrix $U$ contains a $d$-dimensional representation of the climber in the corresponding row of $Y$. We call this representation the \emph{climber embedding}. Similarly, each column of the matrix $V$ contains a $d$-dimensional representation of the problem in the corresponding column of $Y$. We call this representation the \emph{problem embedding}. The dot product of a given climber's embedding and a given problem's embedding is passed through a sigmoid activation function to represent our predicted probability that this climber successfully completes the problem.

The matrices $U$ and $V$ are trained via random initialization followed by stochastic gradient descent. We optimize the matrices to maximize the log likelihood of the observed results under the predicted success probabilities. This is equivalent to minimizing the log loss, expressed as:
\begin{equation}
    \mathscr{L}(Y, U, V) = \sum_{i = 1}^m \sum_{j = 1}^n \, \textrm{LogLoss}(Y_{ij}, \, \sigma( U_{i\cdot} \cdot V_{\cdot j}))
\end{equation}
where the log loss is defined as:
\begin{equation}
    \textrm{LogLoss}(y, \, \hat{y}) = \textrm{BCE}(y, \, \hat{y}) = y \, \log{\hat{y}} + (1-y) \, \log {1 - \hat{y}}
    \label{eq:binary-cross-entropy}
\end{equation}

Note that since this is a binary classification, the log loss in this case is equivalent to the binary cross-entropy (BCE) loss. We optimize this loss function for 1000 epochs using the Adam optimizer \citep{Adam} and experiment with the number of latent factors, $d \in \{1, 2, 3, 4, 5\}$. Note that the case of $d=1$ is very similar to the logistic regression except for the inclusion of problem-specific coefficients. As in the logistic regression, any climbers with fewer than $N$ problems attempted are grouped together as a replacement-level climber. Furthermore, any problems attempted by fewer than 10 climbers are grouped together as a single problem to prevent overfitting.

\subsection{Evaluation Metrics}

To evaluate the predictive performance of our models, we consider several different metrics. Firstly, we use the number of true positives TP, true negatives TN, false positives FP, and false negatives FN to calculate accuracy and F1 score as follows:
\begin{align}
     \textrm{Accuracy} &= \frac{TP+TN}{TP+TN+FP+FN} \\[10pt]
   \textrm{F1 Score} &= \frac{2*TP}{2*TP+FP+FN}
\end{align}
For these two metrics, a higher score indicates better performance. Secondly, we use the predicted probabilities $\hat{y}_{ij}$ to compute the Brier score and log loss as follows:
\begin{align}
    \textrm{Brier Score} &= \frac{1}{\abs{D}} \sum_{(i, j) \in D} (\hat{y}_{ij} - y_{ij})^2 \\[10pt]
    \textrm{Log Loss} &= -\frac{1}{\abs{D}}\sum_{(i, j) \in D}{\textrm{BCE}(y_{ij}, \, \hat{y}_{ij})}
\end{align}
Here, $D$ is the dataset of observed problem attempts and the binary cross-entropy function BCE is as defined in Equation \ref{eq:binary-cross-entropy}. For these two metrics, a lower value indicates better performance. Lastly, we calculate the ROC score as the area under the receiver-operator curve of true positive rate against false positive rate. For this metric, a higher values indicates better performance.

\section{Results}

\subsection{Model Predictive Performance Evaluation}

\begin{figure}[h!]
    \centering
    \includegraphics[width=\textwidth]{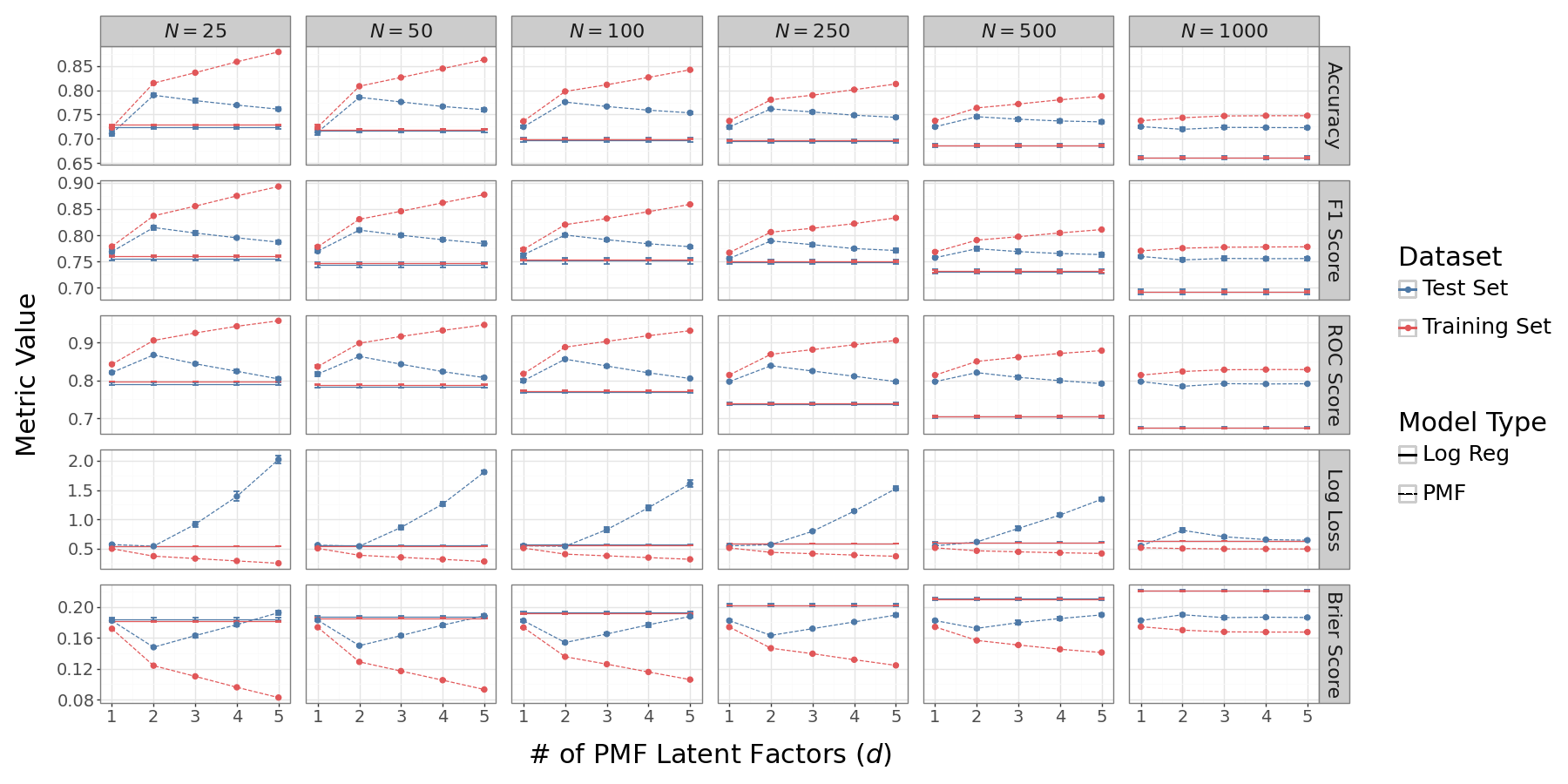}
    \caption{Performance comparison of logistic regression (Log Reg) and PMF models. Error bars (often too small to be visible) indicate 95\% confidence intervals estimated using 5-fold cross-validation. Note that the logistic regression results are plotted as horizontal lines since these models do not use latent factors. The plots are faceted by the replacement level $N$.}
    \label{results}
\end{figure}

In Figure \ref{results}, we present the performance of the logistic regression and PMF models for various hyperparameters. We see that the PMF models generally outperform the logistic regression model among different replacement levels $N$ for both the training and test sets, indicating that the added flexibility of this approach has improved predictive power.

We also see that while the training set performance generally improves for higher number of PMF latent factors $d$, the test performance generally worsens for $d > 2$. This is an indication that the more complex model is overfitting by memorizing  patterns of individual climbers in the training set that do not generalize to the test set. However, the effect of overfitting diminishes for higher replacement levels $N$, where more data is available per climber.
\newpage

For low replacement levels, the PMF models with $d=2$ latent factors exhibit the best performance on the test sets. The fact that these models outperform the $d=1$ models suggest that there is indeed a valuable signal captured by using multi-dimensional representations of climber skill sets. In fact, for $N > 1000$, the model performance is stable enough with enough climber data that an even higher number of latent factors might by appropriate. This result offers evidence that with sufficient data, a nuanced multi-dimensional understanding of climber skill sets can be learned effectively using PMF.

\subsection{Analysis of PMF Climber Embeddings}

We focus on the case of $N=100$ and $d=2$ because it demonstrates a strong test performance while also consisting of a moderate sample size of 216 climbers. To interpret the PMF-generated embeddings, we apply a principal component analysis (PCA) to help explain the variances of the climber embeddings.

\begin{figure}[h!]
    \centering
    \includegraphics[width=1\textwidth,height=1\textheight,keepaspectratio]{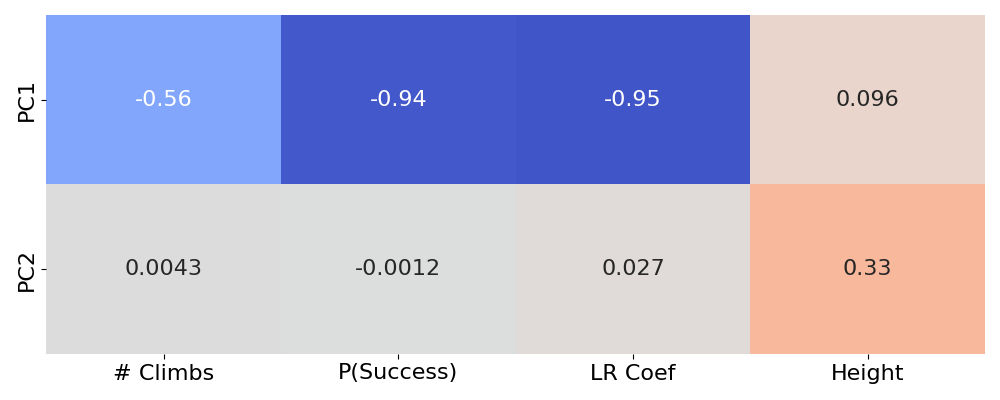}
    \caption{Correlation matrix for climber embedding principal components (for PMF with $N=100$ and $d=2$) and selected variables.}
    \label{climber_emb_corr}
\end{figure}

In Figure \ref{climber_emb_corr}, we show the correlations between the principal components of the embeddings and the following metrics:
\newpage

\begin{itemize}
    \item \textbf{LR Coef}: The logistic regression coefficient for the respective climber
    \item \textbf{\# of Climbs}: The number of problem attempts for that climber in our dataset
    \item \textbf{P(Success)}: The climber's observed average probability of success in our dataset
    \item \textbf{Height}: The climber's height, if available
\end{itemize}

Unsurprisingly, the first principal component of variation in the climber embeddings (PC1) is very strongly correlated with both the logistic regression coefficients ($|\rho| = 0.95$) and climber success rate ($|\rho| = 0.94$). This result suggests that PC1 is a good indicator of the overall climber strength. We also observe a strong correlation between PC1 and the number of problems attempted ($|\rho| = 0.56$), a proxy for professional experience. In bouldering, as in many sports, professional experience is a significant predictor of an athlete’s ability, though it is rarely the sole determinant.

The negligible correlation between PC1 and climber height indicates that height, on its own, is not a strong predictor for overall success in bouldering. However, the moderate correlation between PC2 and climber height shows this component partially picks up on variations with respect to climber's physical characteristics. This is intuitive because different types of boulder problems may suit climbers of different physical traits. For example, boulder problems with big dynamic movements may suit taller, more powerful climbers. Therefore, we hypothesize that the second principal component of variation in the embeddings is capturing the preferred style of climbers and separating climbers of similar abilities based on the types of problems they each perform best at.

\begin{figure}[h!]
    \centering
    \includegraphics[width=1\textwidth,height=1\textheight,keepaspectratio]{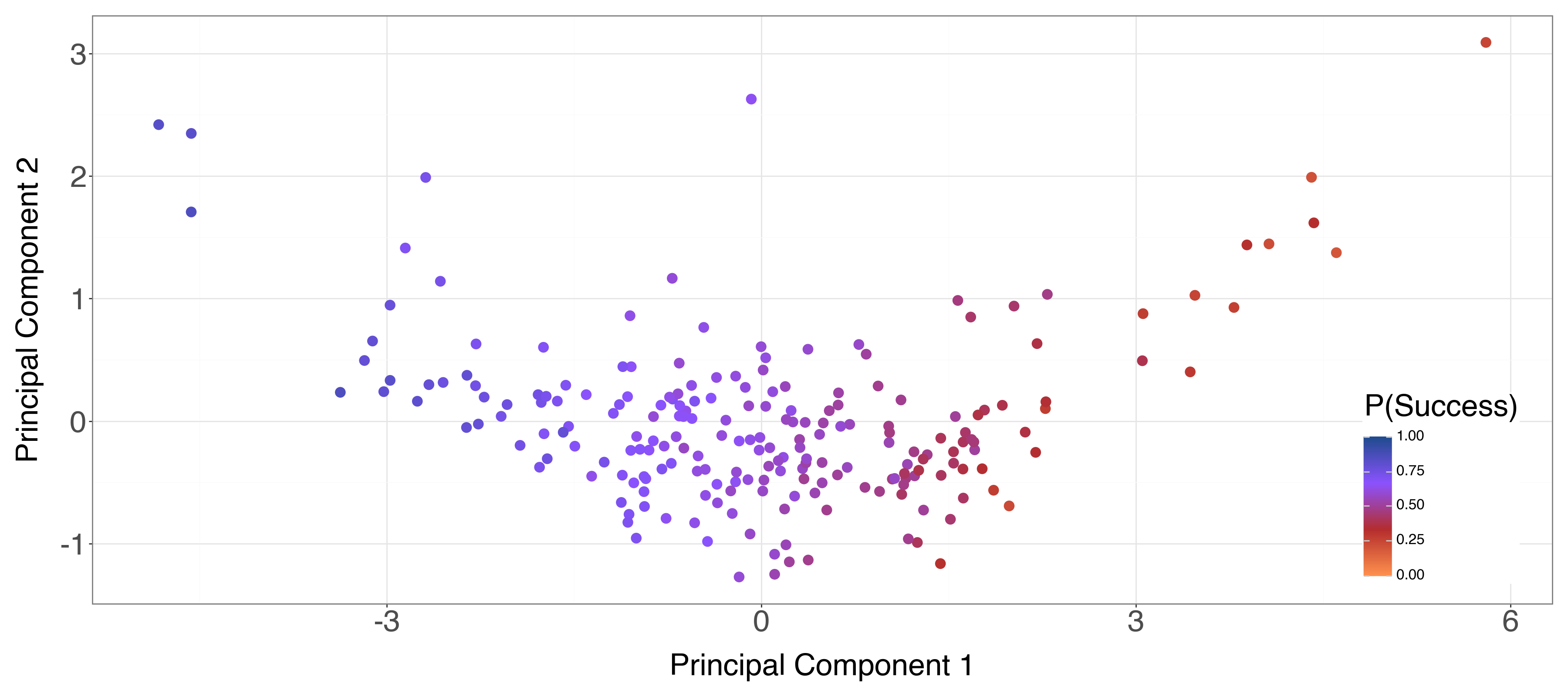}
    \caption{Climber embeddings colored by success rate.}
    \label{climber_success}
\end{figure}

\subsection{Analysis of PMF Problem Embeddings}

\begin{figure}[h!]
    \centering
    \begin{subfigure}[b]{0.48\textwidth}
        \centering
        \includegraphics[width=\textwidth,height=1\textheight,keepaspectratio]{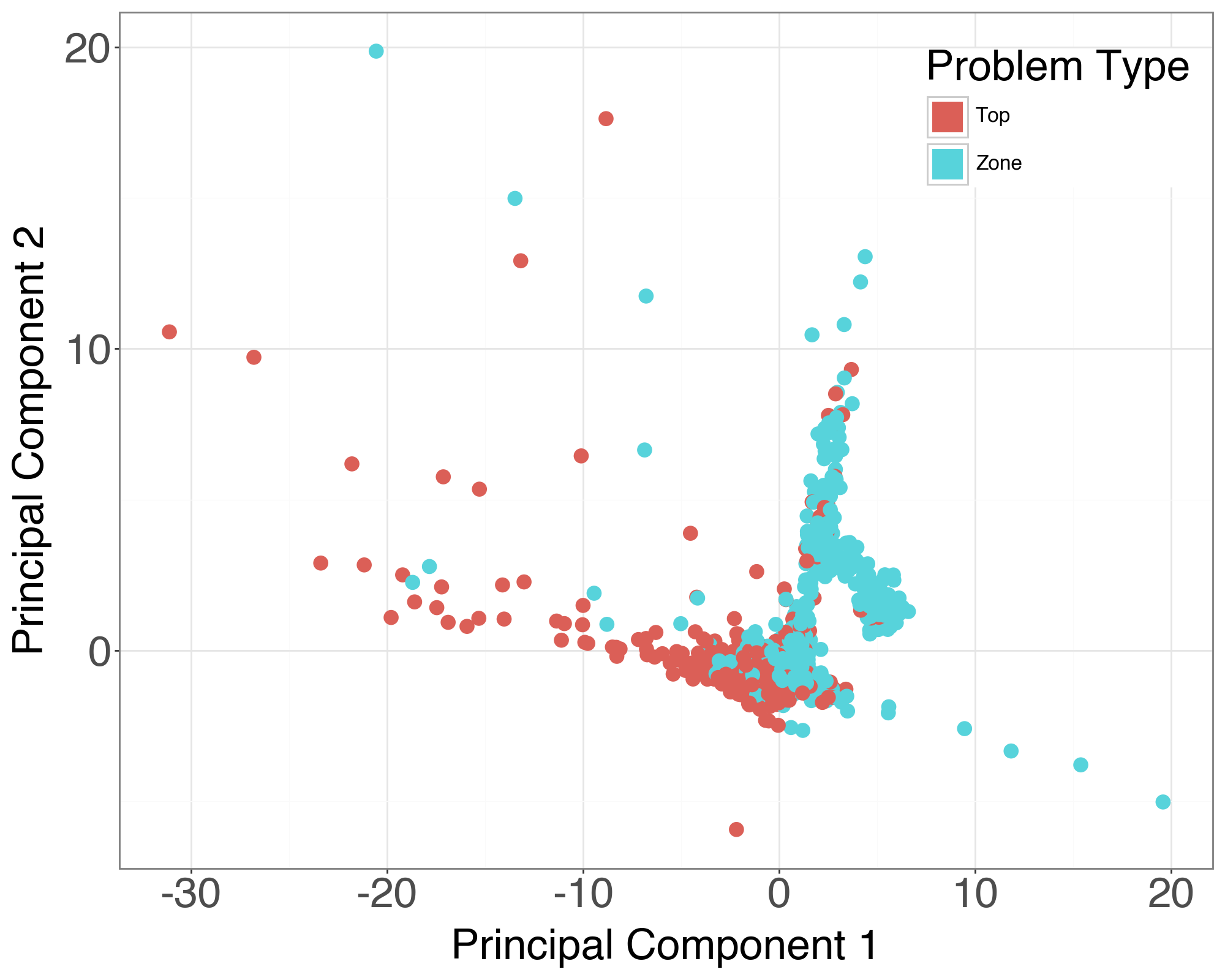}
        \label{category}
    \end{subfigure}
    \hfill
    \begin{subfigure}[b]{0.48\textwidth}
        \centering
        \includegraphics[width=\textwidth,height=1\textheight,keepaspectratio]{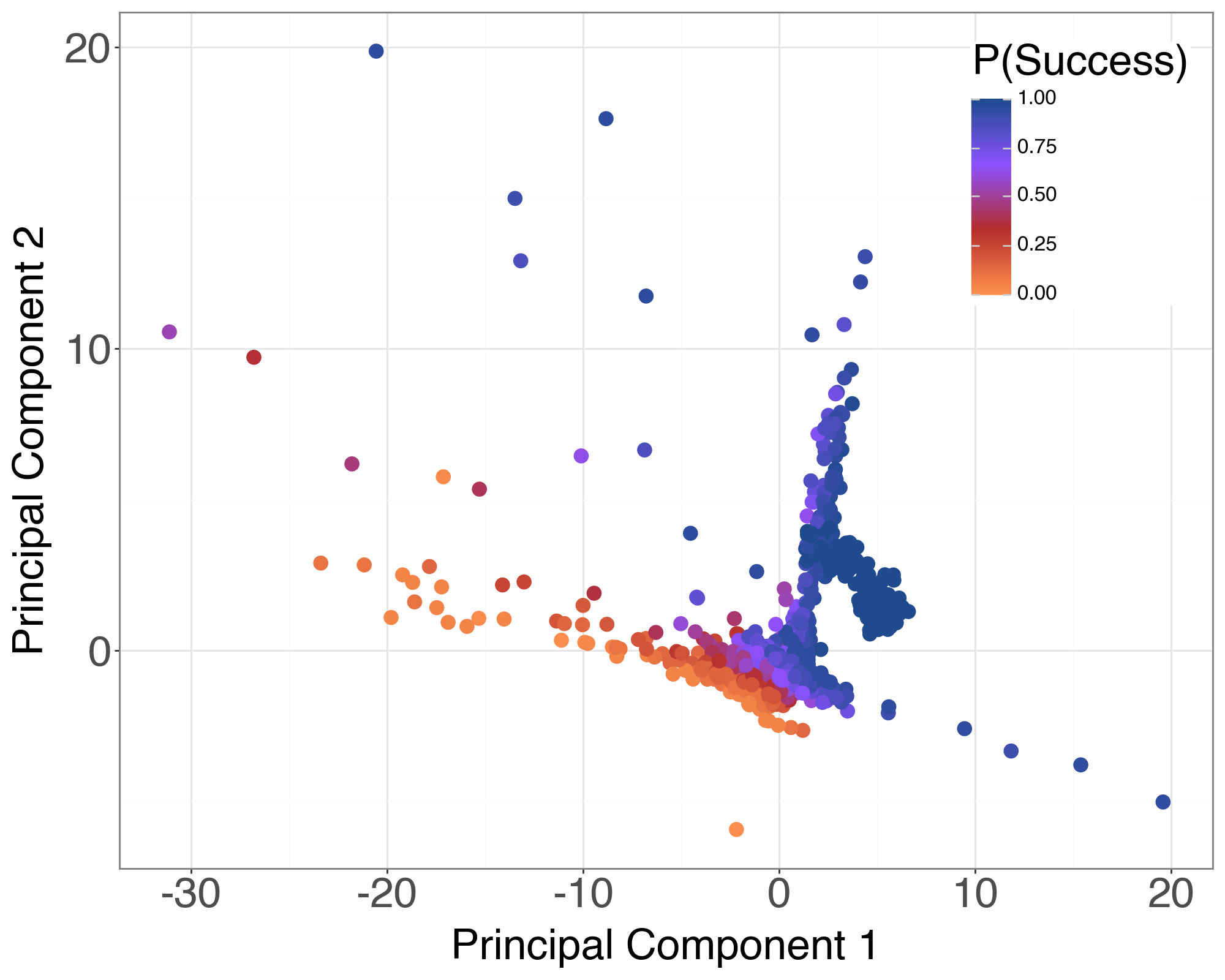}
        \label{success}
    \end{subfigure}
    \caption{Problem embeddings colored by problem type (a) and success rate (b).}
    \label{combined_results}
\end{figure}

As with the climber embeddings, we focus on the case of $N=100$ and $d=2$ and use PCA to interpret the variance in the PMF-generated problem embeddings. We visualize these embeddings in Figure \ref{combined_results} with respect to problem type (zone/top) and difficulty (as measured by the proportion of climbers who succeeded at the problem). Clearly, the PMF embeddings have learned to distinguish between problems based on their difficulty, including separating zones from tops. We hypothesize that both principal components capture aspects of the problem's unique layout that independently contribute to its difficulty.

\section{Future Directions}

There are many interesting directions for future work based on our results. Firstly, rather than viewing success on a problem as binary, leveraging the information about the number of attempts until a climber succeeds at a problem via a survival analysis framework could further refine the model by differentiating between climbers who succeed in topping a problem on their first attempt or only after many attempts.

Secondly, additional information about the specific attributes of each problem, such as the category labels collected by \cite{boulder_types}, would allow us to improve our understanding of the PMF problem embeddings. Ideally, we could learn the problem embeddings directly from the problem geometry such as using a computer vision architecture on video feeds of climbers attempting the problem. Similarly, a more comprehensive dataset of climber characteristics, including anthropometrics such as weight, BMI, and hand size, would enable a deeper exploration and understanding of the PMF climber embeddings.

Thirdly, the framework can be made more robust by explicitly accounting for time-varying effects such as aging to capture changes in climber skill sets over time. This would also improve the model's predictive performance on future competitions by using a recent estimate of each climber's skill rather than an aggregate estimate based on their full career.

Finally, it would be interesting to extend our analysis to women's bouldering competitions or to other sport climbing formats such as lead and speed climbing, to see if our findings hold in these other domains. 

\section{Conclusion}

We use logistic regression and probabilistic matrix factorization to model climber performances in professional bouldering competitions. We show that the multi-dimensional representations learned via probabilistic matrix factorization offer improved predictive performance but can also lead to overfitting in some cases. We therefore discuss the selection of hyperparameters such as replacement level $N$ and latent factor dimension $d$ to mitigate this overfitting. Lastly, we show that the climber embeddings learned by the probabilistic matrix factorization capture both an overall notion of climber strength and orthogonal components of climber skill sets associated with their anthropometrics, which could lead to a more nuanced understanding of climbers' specific strengths and weaknesses.

\newpage
\printbibliography
\end{document}